%% file: main.tex
\documentclass[10pt,twocolumn,letterpaper]{article}

\usepackage[pagenumbers]{iccv} %

\input{preamble}

\definecolor{iccvblue}{rgb}{0.21,0.49,0.74}
\usepackage[pagebackref,breaklinks,colorlinks,allcolors=iccvblue]{hyperref}

\title{\methodname: Towards Unified Language-Image Pretraining}

\author{Zineng Tang, Long Lian, Seun Eisape, XuDong Wang,\\ Roei Herzig, Adam Yala, Alane Suhr, Trevor Darrell, David M. Chan\\
University of California, Berkeley\\
}

\begin{document}
\maketitle
\input{sec/0_abstract}

\input{sec/1_intro}

\input{sec/2_background}

\input{sec/3_methods}

\input{sec/4_results}
\input{sec/5_conclusion}

{
    \small
    \bibliographystyle{ieeenat_fullname}
    \bibliography{main}
}

\clearpage
\appendix

\renewcommand{\theequation}{\thesection.\arabic{equation}}
\renewcommand{\thefigure}{\thesection.\arabic{figure}}
\renewcommand{\thetable}{\thesection.\arabic{table}}

\makeatletter
\@addtoreset{equation}{section}
\@addtoreset{figure}{section}
\@addtoreset{table}{section}
\makeatother

\input{sec/6_appendix}

\end{document}

%% file: preamble.tex
\usepackage{xspace}
\usepackage{amsfonts}
\usepackage{listings}
\usepackage{cite}
\usepackage{multirow}
\usepackage{tabularx}
\usepackage{tcolorbox} %
\usepackage{array} %
\usepackage{caption}

\let\cite\citep

\newcommand{\genaugname}{GeCo\xspace}
\newcommand{\methodname}{TULIP\xspace}
\newcommand{\methodnamelong}{\underline{T}owards \underline{U}nified \underline{L}anguage-\underline{I}mage \underline{P}retraining\xspace}

\newcolumntype{C}{>{\centering\arraybackslash}X}

\newcommand{\xrightarrowdbl}[2][]{%
    \leftarrow\mathrel{\mkern-14mu}\xrightarrow[#1]{#2}
}

\renewcommand{\paragraph}[1]{\vspace{0.5em}\noindent\textbf{#1}$\:\:$}

%% file: sec/0_abstract.tex
\begin{abstract}

Despite the recent success of image-text contrastive models like CLIP and SigLIP, these models often struggle with vision-centric tasks that demand high-fidelity image understanding, such as counting, depth estimation, and fine-grained object recognition. 
These models, by performing language alignment, tend to prioritize high-level semantics over visual understanding, weakening their image understanding. On the other hand, vision-focused models are great at processing visual information but struggle to understand language, limiting their flexibility for language-driven tasks.
In this work, we introduce \methodname, an open-source, drop-in replacement for existing CLIP-like models. Our method leverages generative data augmentation, enhanced image-image and text-text contrastive learning, and image/text reconstruction regularization to learn fine-grained visual features while preserving global semantic alignment. Our approach, scaling to over 1B parameters, outperforms existing state-of-the-art (SOTA) models across multiple benchmarks, establishing a new SOTA zero-shot performance on ImageNet-1K, delivering up to a $2\times$ enhancement over SigLIP on RxRx1 in linear probing for few-shot classification, and improving vision-language models, achieving over $3\times$ higher scores than SigLIP on MMVP.
Our code/checkpoints are available at \url{https://tulip-berkeley.github.io}.

\end{abstract}

%% file: sec/1_intro.tex
\section{Introduction}
\label{sec:intro}

Contrastive image-text (CIT) models, including CLIP \cite{radford2021learning}, SigLIP \cite{zhai2023sigmoid}, and ALIGN \cite{jia2021scaling} have demonstrated state-of-the-art performance on high-level vision-language tasks, excelling in various applications such as retrieving images from text and vice versa, performing zero-shot classification, and serving as core components of vision-and-language models. Their success stems from their ability to leverage billion-scale datasets to create a shared embedding space between image and language inputs, where similar concepts are close together and dissimilar ones are far apart.

\begin{figure}
    \centering
    \includegraphics[width=\linewidth]{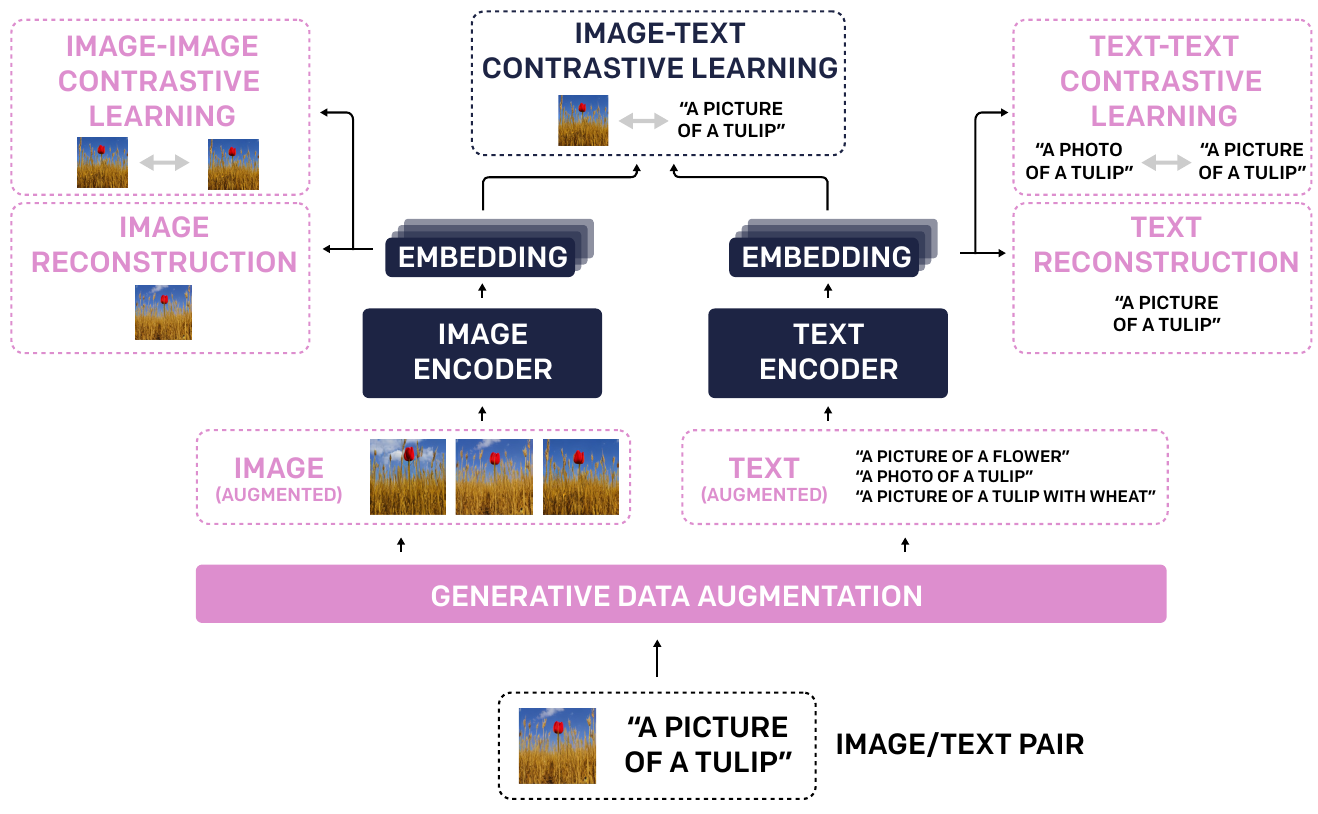}
    \caption{\textbf{TULIP Overview.} Existing contrastive image-text models struggle with high-fidelity visual understanding. \methodname is a drop-in replacement for CLIP which leverages generative data augmentation, global-local patch-wise image contrastive learning, and reconstruction-based feature regularization to learn robust visual features and fine-grained language grounding.}
    \label{fig:teaser}
\end{figure}

Nonetheless, existing CIT approaches come with several notable drawbacks. While representations learned by contrastive image-text models tend to encode the high-level semantics between images and text, encoding global alignment often comes at the cost of reduced performance in visual fine-grained tasks such as spatial reasoning. Existing CIT model representations are thus over-optimized for identifying \textit{what} is present in an image over determining \textit{where} it is located or noticing fine-grained details distinguishing similar objects. This limitation stems from training data and objectives that lack a focus on precise spatial understanding and do not provide the detailed annotations necessary for fine-grained visual differentiation or grounding. As a result, tasks requiring more subtle visual understanding, such as multi-view reasoning, counting, instance segmentation, depth estimation, and object localization, pose a greater challenge compared to high-level tasks.

In this paper, we introduce \methodname (\methodnamelong), an open-source drop-in replacement for existing open-weights CIT models designed to enhance the learning of general-purpose visual features while preserving the language-grounding strengths of current CIT methods. 
Our method addresses two fundamental challenges of existing CIT methods: representation of detailed spatial information, and representation of nuanced visual details. To encode detailed spatial information, we incorporate patch-level global and local multi-crop augmentations and objectives, inspired by methods such as iBOT \cite{zhou2021image} and DINO \cite{oquabdinov2}. To maintain the high-frequency local visual details that image-text contrastive objectives often overlook, we introduce a reconstruction objective. While existing CIT approaches focus on high-level semantic representation and often miss these local details, we find that incorporating them enhances performance in various downstream tasks, such as visual question answering. Finally, we propose a generative data augmentation strategy based on diffusion models, designed to produce challenging hard negatives that refine fine-grained semantic grounding.

We demonstrate the efficacy of \methodname by evaluating it against existing CIT models (such as OpenAI's CLIP \cite{radford2021learning} and the recently introduced SigLIP 2 \cite{tschannen2025siglip}) on a diverse suite of vision-centric downstream tasks covering both traditional and specialized datasets. Specifically, we assess performance on general-purpose zero-shot classification datasets such as ImageNet-1K, iNAT-18, and Cifar-100, as well as fine-grained, task-specific classification datasets including RxRx1, fMoW, and Infographics. We demonstrate that \methodname outperforms SOTA models across all benchmarks (in some cases, even outperforming larger models). We further demonstrate SOTA text-based image retrieval performance in both COCO and Flickr, along with the inverse image to text problem. To examine the model’s robustness in vision-language tasks, we conduct evaluations using our model as a visual encoder for a LLaVA-style model on both the  MMVP and MM-Bench datasets, demonstrating that when used as a drop-in replacement for existing CIT models, \methodname can lead to more than 3x improvements on MMVP (vision-centric downstream tasks) over CIT models without degraded performance on language-centric tasks. Finally, we assess the reasoning and perceptual skills of our approach using the BLINK benchmark, demonstrating up to 12\% relative improvement over SigLIP-trained baselines, and visio-linguistic compositional reasoning skills using the Winoground benchmark where we outperform existing CIT models by up to 30\%, achieving above-random performance in Group-based reasoning—a first for CIT models. 

We summarize our main contributions as follows: (i) We introduce \methodname, a modified image-language pretraining framework that enhances the encoding of fine-grained visual representations while maintaining the language-grounding capabilities of existing ILP methods. (ii) We incorporate patch-level global and local multi-crop augmentations and objectives to improve spatial awareness. (iii) We introduce a reconstruction objective that preserves high-frequency local visual details. (iv) We propose a generative data augmentation strategy based on diffusion models, designed to generate challenging hard negatives that refine fine-grained semantic grounding. (v) We evaluate \methodname on a broad set of vision and vision-language benchmarks, establishing a new state-of-the-art performance in zero-shot classification, fine-grained recognition, object detection, and multi-modal reasoning tasks.

%% file: sec/2_background.tex
\section{Related Work}

\paragraph{Vision-Centric Self-Supervised Learning.}
Vision-centric self-supervised learning has witnessed remarkable progress, driven by the development of learning representations from unlabeled image data. Early approaches like DeepCluster~\cite{caron2018deep} explored clustering-based techniques, while contrastive learning frameworks such as MoCo~\cite{he2020momentum,chen2020improved}, SimCLR~\cite{chen2020simple}, and SwAV~\cite{caron2020unsupervised} leveraged different augmentations of the same image and diverse augmentations to learn powerful representations. Further advancements were made with non-contrastive methods like BYOL~\cite{grill2020bootstrap} and Barlow Twins~\cite{zbontar2021barlow}, which eschewed explicit negative samples. 

More recently, DenseCL~\cite{wang2021dense} introduced dense contrastive learning, and VICReg~\cite{bardes2021vicreg} proposed a variance-invariance-covariance regularization framework. Meanwhile, DINO~\cite{caron2018deep,oquabdinov2} leveraged self-distillation with a momentum encoder, and masked image modeling approaches like MAE~\cite{he2022masked}, DiffMAE~\cite{wei2023diffusion}, and CrossMAE~\cite{fu2024rethinking} demonstrated the effectiveness of reconstructing masked image patches. Finally, I-JEPA~\cite{assran2023self} and V-JEPA~\cite{bardes2024revisiting} advanced self-supervised visual representation learning by introducing masked prediction tasks, which involve predicting abstract representations of masked image regions.

Unlike previous methods, our approach enhances contrastive image-text learning by explicitly incorporating visual local details. This is achieved through patch-level global and local multi-crop augmentations and objectives, complemented by a reconstruction objective, leading to a more comprehensive understanding of visual information.

\paragraph{Generative Data Augmentation.} Generative data augmentation has recently emerged as a powerful technique to expand training datasets beyond traditional transformations. ALIA \cite{dunlap2023diversify} develops a diffusion-based augmentation pipeline that uses language-guided image editing to create realistic domain variations of training images, significantly enhancing dataset diversity and improving classification performance. Similarly, \cite{trabucco2023effective} leverages pre-trained text-to-image diffusion models to perform semantic image edits to create augmented examples, which boosts accuracy in few-shot classification tasks. \cite{shipard2023diversity} proposes a model-agnostic approach using a diffusion model to synthesize class-specific images for unseen categories to improve zero-shot classification performance. \cite{he2022synthetic} showed that such synthetic data can improve model performance in low-data regimes and assist large-scale model pre-training. \cite{azizi2023synthetic} demonstrates that adding diffusion-generated samples to ImageNet yields significant gains in classification accuracy. StableRep \cite{tian2023stablerep} indicates that a model trained exclusively on 20M Stable Diffusion-generated images can learn visual representations rivaling those obtained from training on 50M real images. 

In contrast to prior works that enrich datasets with generative images, our \genaugname integrates generative augmentation directly into the contrastive learning framework. By leveraging large language models to create both positive and negative paraphrases alongside diffusion-based image edits, our dual-modality approach produces richer contrastive views that enhance both visual and textual representations. Unlike methods focused solely on classification or domain shifts, \genaugname refines fine-grained semantic grounding by generating hard negatives that compel the model to discern subtle differences in image-text pairs.

\paragraph{Contrastive Image-Text Learning.} Contrastive image-text (CIT) learning has emerged as a powerful paradigm for learning joint representations of visual and textual information. Through training models on extensive datasets of image-text pairs, CIT enables the alignment of visual and textual representations within a common embedding space. Pioneering works like CLIP~\cite{radford2021learning} and ALIGN~\cite{jia2021scaling} demonstrated impressive zero-shot capabilities and achieved state-of-the-art performance on various vision-language tasks, including image-text retrieval and visual question answering. 

Subsequent efforts have focused on improving the efficiency and scalability of contrastive learning, such as SigLIP~\cite{zhai2023sigmoid}, which introduced a novel sigmoid loss function, and SigLIP 2~\cite{tschannen2025siglip} (introduced concurrent to our work), which extended the training objective of sigmoid loss with additional objectives for improved semantic understanding, localization, and dense features. Despite their successes, these models often struggle with fine-grained visual understanding and tasks requiring precise spatial reasoning. Finally, SLIP~\cite{mu2021slip} also explored the usage of self-supervised learning with language supervision for visual representation learning. However, unlike our approach, SLIP focused solely on image-text and image-image contrastive learning with fixed augmentations.

%% file: sec/3_methods.tex
\begin{figure}
    \centering
    \includegraphics[width=\linewidth]{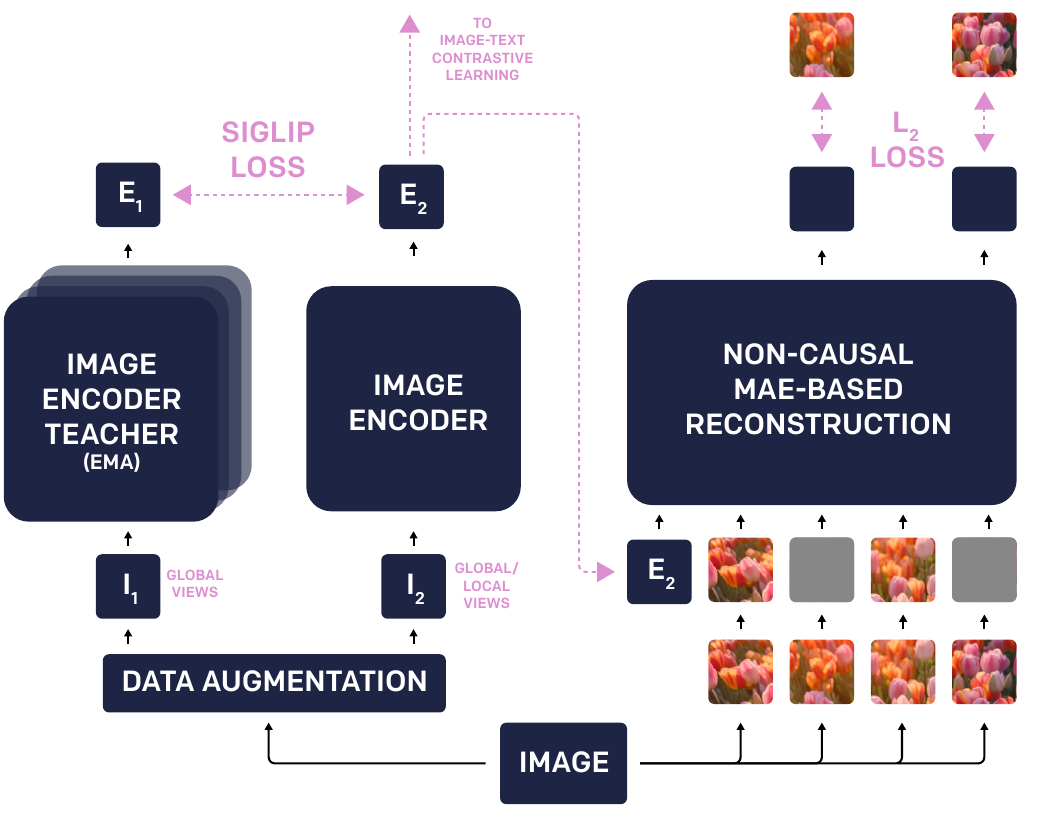}
    \caption{\textbf{TULIP Image Encoder.} Images undergo both traditional augmentations (such as cropping and color jittering) and generative augmentations via \genaugname, which leverages large generative models to create semantically consistent or semantically altered views. These views are then used for image-image and image-text contrastive learning. Additionally, a masked autoencoder (MAE)-based reconstruction loss is applied to encourage the model to encode both semantic and fine-grained details.}
    \label{fig:image_path}
\end{figure}

\section{\methodname}

We introduce \methodname, a high-performing image-text contrastive model that unifies several diverse contrastive learning paradigms to improve representation learning. The underlying insight behind several of the contributions of \methodname is that images and their associated captions represent different ``views'' or perspectives of an underlying ``reality,'' an observation recently explored in \citet{huh2024platonic}. For example, a picture of a cat with a bench, and the caption ``a cat is sitting on a bench'' present different observations of the same underlying true situation. Contrastive learning serves to unify these ``views'' in an unsupervised way - taking several views and projecting them to the same point in a representation latent space. Thus, defining what constitutes a valid view of the underlying content is fundamental in developing a contrastive learning approach. 

In this section, we first discuss how \methodname uses images and text to provide different views in the contrastive learning process (See~\autoref{sec:contrasview}). Next, we present how \methodname creates different views from the ``reality'' with generative augmentations (See \autoref{ssec:generative_transformations}), and how \methodname regularizes the training with reconstruction loss to learn a more robust representation (See~\autoref{sec:reconstruction}).

\subsection{Diversifying Contrastive Views}
\label{sec:contrasview}
Previous image-text contrastive learning approaches primarily contrast an image with its corresponding text, while image-image contrastive learning methods contrast an image with an augmented version of itself. We propose a unification of these approaches by treating every transformation of an image or text as a valid \textit{view} of the underlying semantic content, which is then incorporated into the contrastive learning framework.

Thus, our contrastive learning loss comprises three key components: image-text contrastive learning, image-image contrastive learning, and text-text contrastive learning, as explained in~\cref{fig:teaser}.

The contrastive loss in our method sources from SigLIP~\cite{tschannen2025siglip}. Denote $x$ and $y$ as two views from the same underlying content with batch size $|\mathcal{B}|$:
\begin{equation}
L_\text{SigLIP}(\{\mathbf{x}\}, \{\mathbf{y}\}) = - \frac{1}{|\mathcal{B}|} \sum_{i=1}^{|\mathcal{B}|} \sum_{j=1}^{|\mathcal{B}|} \log \underbrace{\frac{1}{1 + e^{z_{ij} (-t \mathbf{x}_i \cdot \mathbf{y}_j + b)}}}_{\mathcal{L}_{ij}}
\label{eq:siglip_loss}
\end{equation}

\begin{figure}
    \centering
    \includegraphics[width=\linewidth]{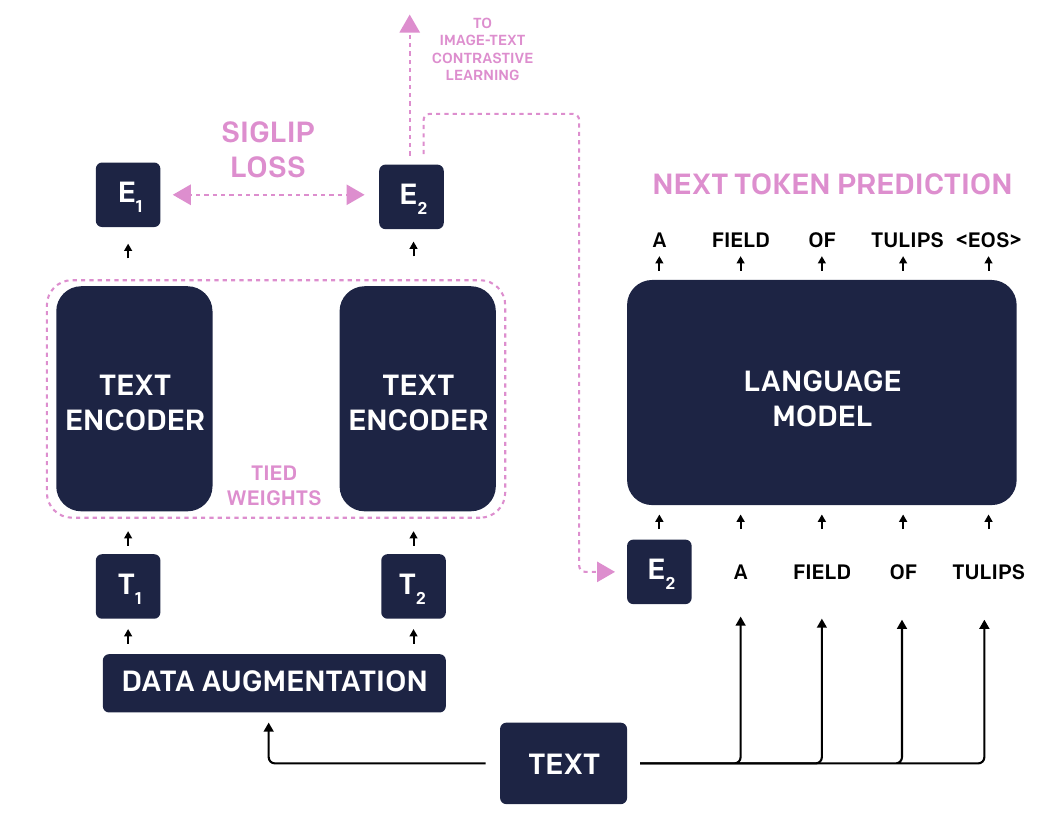}
    \caption{\textbf{TULIP Text Encoder.} Text undergoes generative augmentation through paraphrasing and controlled semantic alterations using large language models, generating both positive and negative contrastive pairs. These pairs are used for both text-text and image-text contrastive learning with a SigLIP objective. Similar to image reconstruction, a causal decoder (based on T5) is used for text reconstruction, ensuring that the model retains both high-level semantics and fine-grained linguistic detail.}
    \label{fig:text_path}
\end{figure}

\paragraph{Image-Text Contrastive Learning.}
For each image $i$ in the batch $\mathcal{B}$, we use the standard image-text contrastive learning objective from SigLIP:

\begin{equation}
L_\text{I-T} = L_\text{SigLIP}(\{\mathbf{x}_I\}, \{\mathbf{x}_T\})
\end{equation}

\paragraph{Image-Image Contrastive Learning.}
To construct transformed images, we leverage a generative model instead of the traditional fixed set of augmentations commonly used in contrastive learning. Our generative transformations significantly outperform standard augmentation techniques such as those used in DINO, leading to more robust representations. We present details of the generative transformations in~\autoref{ssec:generative_transformations}. Given the original image embedding $\mathbf{x}_I$ and the transformed image embedding $\mathbf{x}_I'$, we define our image-image contrastive loss as:

\begin{equation}
L_\text{I-I} = L_\text{SigLIP}(\{\mathbf{x}_I\}, \{\mathbf{x}_I'\})
\end{equation}

\paragraph{Text-Text Contrastive Learning.} To enhance textual representations, we apply generative augmentation using a language model including syntactic paraphrasing and synonym replacement (See  \autoref{ssec:generative_transformations}). Given the original text embedding $\mathbf{x}_T$ and the transformed text embedding $\mathbf{x}_T'$, we define our text-text contrastive loss as:

\begin{equation}
L_\text{T-T} = L_\text{SigLIP}(\{\mathbf{x}_T\}, \{\mathbf{x}_T'\})
\end{equation}

\noindent Our overall contrastive learning loss is as follows:

\begin{equation}
    L_\text{cont} = L_\text{I-T} + L_\text{I-I} + L_\text{T-T}
\end{equation}

\paragraph{Image Encoder.} The image encoder for \methodname is shown in \autoref{fig:image_path}. Following DINOv2, we use an EMA teacher model, combined with local/global view splits (where the teacher only sees global views, and the student sees both global and local views). Similar to DINOv2, we leverage the embeddings generated by the teacher model for image-image contrastive learning and image-text contrastive learning. In our experiments, the image encoder takes the form of a SigLIP image encoder, which is a ViT model \cite{dosovitskiy2020image}. The reconstruction regularization shown in the pathway is discussed in \autoref{sec:reconstruction}.

\paragraph{Text Encoder.} The text encoder for \methodname is shown in \autoref{fig:text_path}. For text encoding, there is no clear global/local structure in the views, so we do not use an EMA teacher and instead leverage a text encoder with directly tied weights. For a text encoder, we use SigLIP's language encoder. The reconstruction regularization is further discussed in \autoref{sec:reconstruction}.

\subsection{\genaugname: Generating Diverse Contrastive Views}
\label{ssec:generative_transformations}

Existing models for contrastive learning focus on using fixed sets of views to force models to learn semantic invariance. While fixing the set of potential views is simple, choosing the right views is a challenging task. The particular set of views chosen can also impact the level of features learned by the model. In DINO, models are trained to match local/small crops of the images with global crops of images, leading to strong global semantic features, but often leading models to ignore complex relationships between objects. Recent work has shown that many generative models inherently encode semantics at natural levels, for example, GPT-4V performs well when measuring semantic distance in natural language \cite{chan2023clair}, and Stable Diffusion latent encode semantic correspondences between images \cite{hedlin2023unsupervised}. This motivates a view generation approach that relies on semantic information encoded by these large generative models, in addition to a base set of simple pixel-level augmentations.

Towards such a generative augmentation, we introduce \genaugname (GEnerative COntrastive view augmentation), a method that leverages large generative models (both language and image), to generate semantically equivalent (and semantically distinct but visually similar) augmentations automatically during training. \genaugname alters both image and text automatically in the axes of perception, space, and time, to create positive and negative pairs that are fed to the contrastive components making up TULIP. \genaugname generates two types of view pairs:
\vspace{0.5em}
\begin{itemize}
    \item \textbf{Positive} views are views of the same content which contain identical semantics viewed in a different (but similar) way. These views should be ``closer'' in semantic space. For example, rotating the camera around an object slightly does not significantly change the semantics of an image, but can change the local pixel values.
    \item \textbf{Negative} views are views of content that are semantically distinct, but contain many similar image characteristics, for example, adding a ``car'' into the image of a ``bike'' creates a new image which is semantically distinct, but contains many of the same visual features.
\end{itemize}
\vspace{0.5em}

\begin{figure}
    \centering
    \includegraphics[width=\linewidth]{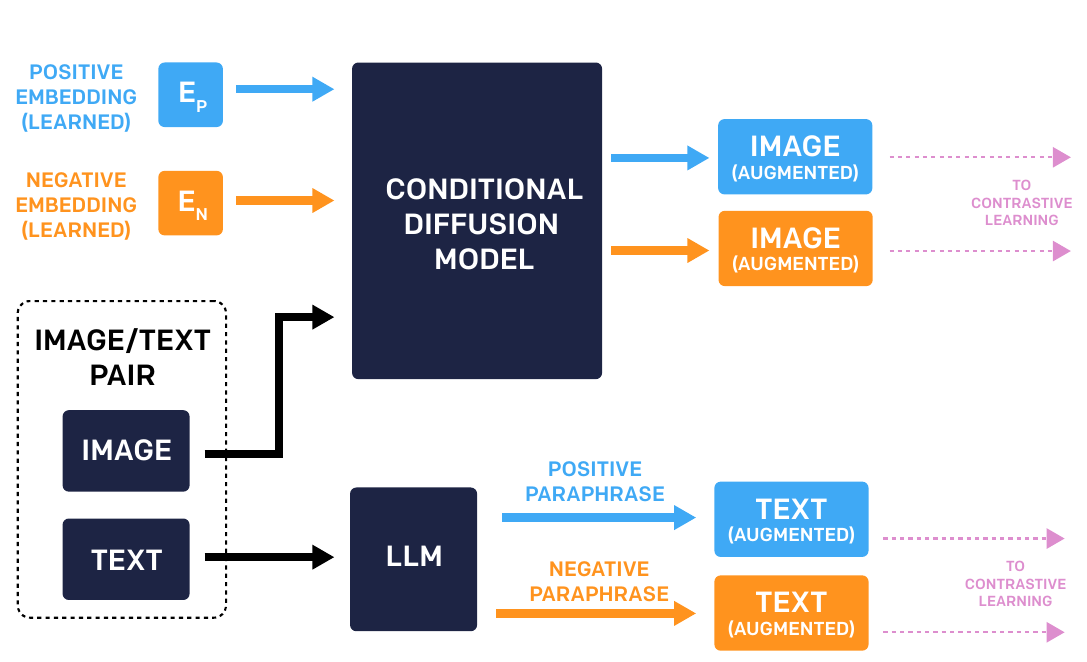}
    \caption{\textbf{Overview of \genaugname.} Our generative augmentation framework leverages large generative models to create diverse contrastive views by generating both positive and negative augmentations for images and text. For text augmentation, we use Llama-3.1-8B-Instruct to generate paraphrases and semantically altered text variations. For image augmentation, we fine-tune an instruction-based image editing model (e.g., InstructPix2Pix) fine-tuned using soft-prompting to generate semantically consistent (positive) and semantically altered (negative) views.}
    \label{fig:gecov}
\end{figure}

Unfortunately, such paired data is often unavailable, thus, \genaugname makes use of generative modeling to generate these positive and negative views from existing pairs of images and text. The general process for \genaugname is shown in \autoref{fig:gecov}, and consists of two components: language augmentation and image augmentation.

\paragraph{Language Augmentation.} To augment language, several methods (primarily targeted at hallucination reduction) have pursued random word deletion or word synonym replacement \cite{shekhar2017foil}. Here, we leverage a large language model (Llama-3.1-8B-Instruct) to perform similar-style augmentations. We ask the model to directly paraphrase the content of the text to produce positive (where the semantics are identical) and negative paraphrases (where the semantics have been subtly altered). By relying on language models to make this decision, we can take advantage of the underlying semantic understanding in the LLM, and avoid pre-defining a specific level of semantic similarity. The prompts, given in \autoref{app:gecov}, differ for positive and negative augmentation. When generating positive samples, concretely, the LLM should not change the semantics such as objects, counting, layout, etc, while it can paraphrase text by syntactics, synonyms, etc. When generating negative samples, we can follow similar logic to change the semantics of the text, such as changing ``5 apples'' $\rightarrow$ ``4 apples'' or changing the compositional components of the image such as ``chair to the left of table'' $\rightarrow$ `table to the left of the chair.

\paragraph{Image Augmentation.} To augment the images, we fine-tune (using soft-prompting) an instruction-based image editing generative model to generate both positive and negative augmentations of an image. Formally, for an image-editing model $G(I,E)$ where $I$ is an image and $e$ is a vector embedding, we learn embeddings $E_p$ (positive) and $E_n$ (negative) corresponding to positive and negative views. To train these embeddings, we draw on several ``natural'' sources of image augmentation. In addition to traditional image augmentations (i.e., simple color jitter, flipping, global cropping, Gaussian blur, etc.) we also consider several further augmentations. For positive training, the primary addition is video data, where we consider closely related frames ($< 0.2s$ apart) to be semantically identical, and multi-view data, where we consider multiple views of the same object to be semantically identical. For negative training, we use large-scale datasets for semantic image editing, as each image edit encodes the image's semantic transformation.

\begin{figure}
    \centering
    \includegraphics[width=\linewidth]{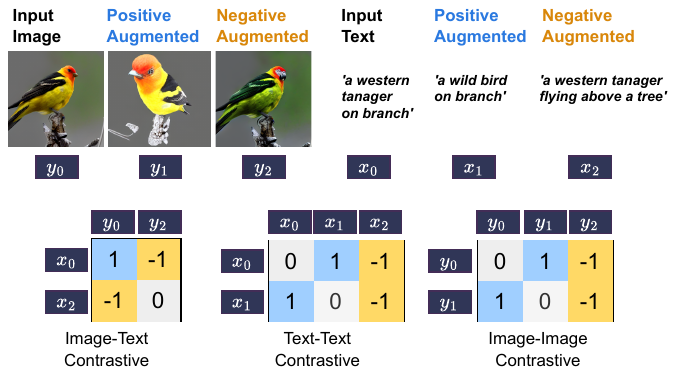}
    \caption{(Top) \genaugname generates positive and negative augmentations of both images and text, (Bottom) \methodname uses these augmentations during training time with corresponding weights (+1 for positive pair, -1 for negative pair, 0 to ignore).}
    \label{fig:gecov_example}
\end{figure}

Together, \methodname supports taking an image and paired text, and generating augmented positive and negative views. We can then use these views for training, either online during training time inference or by caching the augmentations and re-using them during the training process as shown in \autoref{fig:gecov_example}. More formally, in the case of image–image or text–text contrastive learning, \genaugname takes an input (image or text) and produces both an augmented positive view and an augmented negative view. Following the notation from \autoref{sec:contrasview} (with loss \( L_{\text{SigLIP}} \)), let $x = \{x_1,\dots,x_i,\dots,x_n\}$ be input images (or texts) and let $y = \{y_1,\dots,y_i,\dots,y_n\}$ be positive and negative augmented views of that image (or text). Define \(\mathcal{N}\) as the set of indices corresponding to negative views. In \autoref{eq:siglip_loss}, we then set:
\begin{equation}
z_{ij} = -\mathbf{1}\{j\in\mathcal{N}\}
\end{equation}
meaning that \(z_{ij} = -1\) (two elements are a negative pair) whenever the \(j\)th view is negative. In image–text contrastive learning, let $y = \{y_1,\dots,y_j,\dots,y_n\}$ be the generated augmented texts. \genaugname generates \textit{only} negative augmented views for both the image and the text, and we set:
\begin{equation}
z_{ij} = -1 \quad \text{if } i\in\mathcal{N} \oplus j\in\mathcal{N}
\end{equation}
Note that this omits the computation for pairs where both $i$ and $j$ belong to $\mathcal{N}$ (since it's unknown what their correspondence is), and focuses on situations where we know the image or text does not match the true value.

\subsection{Regularization with Reconstruction}
\label{sec:reconstruction}

While incorporating a wide range of contrastive views through generative augmentation alone can help to improve the performance of the model on fine-grained semantics, this process also introduces hidden invariance in our model, where the different augmentations of an image encode to the same point. While such invariance is helpful for representation learning, it often leads to reduced performance on high-fidelity visual-centric tasks (such as color-identification, orientation, or depth-estimation). To encourage the model to balance this high-frequency information with the representation of semantics, we additionally add a pixel-level reconstruction objective to the latent vector of the model. The underlying assumption is that if the model can encode the information necessary for reconstructing the image itself from the latent space, it will also encode key visual detail (such as color/texture) while remaining invariant in the semantic space (due to the contrastive objective). 

The reconstruction objectives are shown in \autoref{fig:image_path} for the image pathway, and \autoref{fig:text_path} for the text pathway. For image reconstruction, we leverage a masked autoencoder (MAE) style model augmented with the embedding as a ``bottleneck'' for the information. Using MAE encourages the model to encode shape information and high-entropy detail instead of global patterns (as those global patterns can easily be inferred from unmasked patches). For the text model, we leverage a causal decoder (based on T5), with the text embedding as the initial text token. The loss from the regularization is formatted as:
\begin{equation}
    L_\text{recons} = \lambda_{i}L_\text{image-recons} + \lambda_{t}L_\text{text-recons}
\end{equation}
where $\lambda_i$ and $\lambda_{t}$ represent a weighting tradeoff between the reconstruction loss and other objectives in our network. Since reconstruction can be expensive during training, to ensure minimal computational overhead we compute reconstruction in both modalities, but using the latent vectors for \textit{only one} of the two modalities during each pass. For example, in image-image contrastive learning, we compute the reconstruction loss from one of the image embeddings, and later in the image-text contrastive learning, text reconstruction loss is \textit{also} computed from the pre-existing image embedding (this is reasonable, as the contrastive objectives encourage the vectors coming from each positive pair to be the same at convergence).

Overall, \methodname is pre-trained in one pass with a weighted combination of losses:
\begin{equation}
    L_\text{\methodname} = \lambda_c L_\text{cont} + \lambda_r L_\text{recons}
\end{equation}

%% file: sec/4_results.tex
\section{Experiments \& Results}

\input{sec/tables/zero_shot}
\input{sec/tables/linear_probing}
\input{sec/tables/winnoground}
\input{sec/tables/blink}
\input{sec/tables/mmvp}

In this section, we discuss the experimental design, training procedure, and experimental results for \methodname.

\subsection{Experimental Design}

\paragraph{Data.} To train \genaugname, as described in \autoref{ssec:generative_transformations}, we use video and multi-view datasets for our diffusion model. For next-frame prediction, we sample consecutive frames (within 0.2 seconds) from the WebVid-10M dataset \cite{Bain21}. For multi-view prediction, we use MVImgNet \cite{yu2023mvimgnet}, and for negative view generation, we incorporate datasets from InstructPix2Pix \cite{brooks2022instructpix2pix}. To paraphrase text for augmentation, we leverage the Llama-3.1-8B-Instruct model \cite{dubey2024llama}.

For model pre-training, we train all variants of \methodname using 500M samples from the DataComp-1B dataset \cite{gadre2023datacomp}. To augment the data, we randomly replace 20\% of the original captions with re-captioned data from \citet{li2024if}. During text reconstruction, we find that increasing the proportion of re-captioned data improves results, so we replace 50\% of the base captions with re-captioned data.

\paragraph{Model} The base model architecture and structure are analogous to SigLIP \cite{zhai2023sigmoid}, and we initialize all model variant weights from their respective SigLIP models. Additional components, including projection layers for image-image and text-text contrastive learning, as well as image and text decoders, are added on top of the pretrained SigLIP visual and text backbones and trained from scratch. 

\paragraph{Optimization.} We use Adam optimizer with learning rate $10^{-5}$, weight decay $10^{-4}$, and gradient clipping to norm 2. We set the batch size to 49,152. The image-text, text-text, image-image contrastive learning, and reconstruction are conducted in the same optimization step for efficiency. Our models are trained with up to 32 A100 GPUs over the course of several days. 

\subsection{Vision-Language Understanding}

Our first experiments focus on evaluating the quality of the image-text representations learned by \methodname, where we explore zero-shot classification, text-to-image and image-to-text retrieval, and linear probing for fine-grained classification datasets.

\paragraph{Zero-Shot Classification.} We first benchmark \methodname on zero-shot classification (ImageNet \cite{deng2009imagenet} (1-shot/10-shot), ImageNet v2 \cite{recht2019imagenet}, ImageNet ReaL \cite{beyer2020we} and ObjectNet \cite{barbu2019objectnet}) following the general protocol from \citet{zhai2023sigmoid}, with results in \autoref{tab:so14_g16}. Generally, \methodname outperforms existing approaches within their parameter classes, and represents significant improvements over existing open-source models such as OpenCLIP.

\paragraph{Text-To-Image Retrieval.} In addition to zero-shot classification, we also benchmark on image-retrieval benchmarks (both text-to-image and image-to-text using the COCO \cite{lin2014microsoft} and Flickr-30K \cite{plummer2015flickr30k} datasets), where \methodname significantly outperforms existing benchmark models, particularly in text-to-image modeling at larger scales. 

\paragraph{Linear Probing.} While \methodname performs well on large-scale object understanding benchmarks, many of the improvements that we target in this work are focused on understanding fine-grained detail. Towards this end, we explore the performance of \methodname when training linear probes on domain-specific data. Towards understanding such performance, we evaluate on the IN-1K~\cite{deng2009imagenet}, iNAT-18 \cite{van2018inaturalist}, CIFAR-100 \cite{krizhevsky2009learning}, RxRx1 \cite{sypetkowski2023rxrx1}, fMoW \cite{christie2018functional}, and Infographic \cite{mathew2022infographicvqa} datasets (see \autoref{app:datasets} for detailed dataset descriptions). The results, shown in \autoref{tab:lienar_probe} show that \methodname clearly outperforms existing vision and language representations for fine-grained/detail-oriented tasks (for example, achieving almost twice the performance of SigLIP on RxRx1, and higher performance than DINOv2 alone), while maintaining high-quality language representations (achieving 24\% relative improvement over DINOv2 and outperforming SigLIP on the Infographic dataset). 

\paragraph{Compositional Reasoning.} To evaluate \methodname's ability to understand the composition of images, we further evaluate on the Winnoground dataset \cite{thrush2022winoground}. The results are shown in \autoref{tab:winnoground}, and clearly demonstrate that \methodname is able to perform visual reasoning at a high level compared to existing vision and language models. 

\subsection{Vision \& Language Models}

One of the motivations for developing strong vision and language models is their applications as feature-encoders for large-scale multimodal models such as LLaVA~\cite{liu2023visual, liu2024improved}. To evaluate our model's performance in these applications, we fine-tune Llama-3.2 11B using a set of visual encoders using the LLaVA mixture data. We then evaluate their performance on several benchmarks, including the BLINK benchmark \cite{fu2024blink} (which consists of 14 primarily perceptual tasks including correspondence, visual similarity, and depth estimation), the MMVP benchmark \cite{tong2024eyes} (which tests a model's visual capability), and LLaVA Bench \cite{liu2023visual} (which tests a model's ability to perform conversation, detail description and complex reasoning).

Results on the BLINK dataset are shown in \autoref{tab:blink}. We can see here that \methodname performs strongly across all classes of problems, performing particularly well in vision-driven tasks compared to base methods, where \methodname outperforms GPT-4o in tasks such as spatial reasoning and localization.

The results on MMVP and LLaVA are shown in \autoref{tab:vllm}. While DINOv2-fine-tuned models perform well on the MMVP benchmark, they struggle with language-centric tasks, while CLIP-style models perform better on language-centric tasks, but struggle with visual perception. \methodname allows the best of both worlds in a single model, outperforming DINOv2 and SigLIP in their respective best tasks. 

\paragraph{Ablations.} \autoref{tab:vllm} also shows the performance of \methodname with several components removed. We can see that the largest improvements on MMVP are drawn from the image-image contrastive learning, along with our base data training pipeline. Reconstruction serves to further improve both the vision and LLaVA benchmark performance. \genaugname primarily improves the performance on vision-centric tasks. Interestingly, the LLaVA bench performance seems saturated (in regards to both scale and improvement), suggesting that improving performance on this task requires improvements in the large language model or visual adapter.

%% file: sec/tables/zero_shot.tex
\begin{table*}[t]
\centering
\small
\caption{Zero-shot classification results (\% accuracy) on ImageNet-1K (val, v2, ReaL, 10-shot), ObjectNet, and text-image/image-text retrieval for \methodname vs. several existing SOTA vision and language models.}
\label{tab:so14_g16}
\begin{tabularx}{\linewidth}{llcXccccccccc}
\toprule
\multirow{2}{*}{Model} & 
\multirow{2}{*}{Method} & 
\multirow{2}{*}{Res.} & 
\multirow{2}{*}{Seq.} & 
\multicolumn{5}{c}{\textbf{Classification}} & 
\multicolumn{2}{c}{\textbf{COCO}} & 
\multicolumn{2}{c}{\textbf{Flickr}} \\
\cmidrule(lr){5-9}\cmidrule(lr){10-11}\cmidrule(lr){12-13}
 & & & & IN-val & IN-v2 & IN-ReaL & ObjNet & IN-10s & T$\to$I & I$\to$T & T$\to$I & I$\to$T\\
\midrule
\multirow{7}{*}{B/16}
  & OpenAI CLIP & 224 & 196 & 68.3 & 61.9 & – & 55.3 & – & 33.1 & 52.4 & 62.1 & 81.9 \\
  & Open CLIP & 224 & 196 & 70.2 & 62.3 & – & 56.0 & – & 42.3 & 59.4 & 69.8 & 86.3 \\
  & MetaCLIP & 224 & 196 & 72.4 & 65.1 & – & 60.0 & – & 48.9 & – & 77.1 & – \\
  & EVA CLIP & 224 & 196 & 74.7 & 67.0 & – & 62.3 & – & 42.2 & 58.7 & 71.2 & 85.7 \\
  & DFN & 224 & 196 & 76.2 & 68.2 & – & 63.2 & – & 51.9 & – & 77.3 & – \\
  & SigLIP & 224 & 196 & 76.2 & 69.5 & 82.8 & 70.7 & 69.9 & 47.2 & 64.5 & 77.9 & 89.6 \\
  & SigLIP 2 & 224 & 196 & 78.2 & 71.4 & 84.8 & 73.6 & 72.1 & 52.1 & 68.9 & 80.7 &  93.0 \\
  & \methodname & 224 & 196 & \textbf{79.5} & \textbf{73.0} & \textbf{86.2} & \textbf{74.2} & \textbf{73.8} & \textbf{54.2} & \textbf{70.1} & \textbf{81.8} & \textbf{93.9} \\
\midrule
\multirow{5}{*}{So/14}
  & \multirow{2}{*}{SigLIP} 
    & 224 & 256 & 82.2 & 76.0 & 87.1 & 80.5 & 78.2 & 50.8 & 69.0 & 76.6 & 90.7\\
  & & 384 & 729 & 83.2 & 77.1 & 87.5 & 82.9 & 79.4 & 52.0 & 70.2 & 80.5 & 93.5\\
  & \multirow{2}{*}{SigLIP 2} 
    & 224 & 256 & 83.2 & 77.7 & 87.8 & 84.6 & 79.5 & 55.1 & 71.5 & 84.3 & 94.6 \\
  & & 384 & 729 & 84.1 & 78.7 & 88.1 & 86.0 & 80.4 & 55.8 & 71.7 & \textbf{85.7} & 94.9  \\
  & \methodname
  & 384 & 729 & \textbf{85.0} & \textbf{79.5} & \textbf{89.0} & \textbf{87.2} & \textbf{80.9} & \textbf{56.3} & \textbf{72.0} & 85.3 & \textbf{95.1} \\
\midrule
\multirow{3}{*}{g/16} 
  & \multirow{2}{*}{SigLIP 2} 
    & 256 & 256 & 84.5 & 79.2 & 88.3 & 87.1 & 82.1 & 55.7 & 72.5 & 85.3 & 95.3 \\
  & & 384 & 576 & 85.0 & 79.8 & 88.5 & 88.0 & 82.5 & 56.1 & 72.8 & 86.0 & 95.4 \\
  & \methodname
  & 384 & 576 & \textbf{85.3} & \textbf{80.0} & \textbf{89.6} & \textbf{88.6} & \textbf{82.9} & \textbf{57.8} & \textbf{73.0} & \textbf{87.2} & \textbf{95.7}\\
\bottomrule
\end{tabularx}
\end{table*}

%% file: sec/tables/linear_probing.tex
\begin{table}[t]
\centering
\scriptsize
\caption{Results (\% accuracy) of a linear probe applied to representations learned by existing representation models. \methodname performs strongly across all datasets, even outperforming significantly larger vision foundation models such as AIMv2 3B.}
\label{tab:lienar_probe}
\begin{tabularx}{\linewidth}{lXXXXXX}
\toprule
\textbf{Model} & \textbf{IN-1k} & \textbf{iNAT-18} & \textbf{Cifar 100} & \textbf{RxRx1} & \textbf{fMoW} & \textbf{Info} \\
\toprule
MAE & 82.2 & 70.8 & 87.3 & 7.3 & 60.1 & 50.2 \\
DINOv2 (L/16) & 87.2 & 83.0 & 95.6 & 9.0 & 65.5 & 59.4 \\
OAI CLIP (B/16) & 85.7 & 73.5 & 89.7 & 5.7 & 62.0 & 66.9 \\
FN-CLIP & 86.9 & 76.4 & 93.9 & 6.1 & 63.4 & 68.1 \\
SigLIP (So/14) & 87.3 & 77.4 & 91.2 & 4.6 & 64.4 & 72.3 \\
AIMv2 (H/14) & 87.5 & 77.9 & 93.5 & 5.8 & 62.2 & 70.4 \\
AIMv2 (3B,448px) & 89.5 & \textbf{85.9} & 94.5 & 9.5 & 66.1 & \textbf{74.8} \\
\midrule
\methodname (B/16) & 85.9 & 81.2 & 93.9 & 7.4 & 63.0 & 69.8 \\
\methodname (So/14, 384) & 89.0 & 84.2 & 96.4 & 9.3 & 65.8 & 73.7 \\
\methodname (g/16, 384) & \textbf{89.6} & 85.8 & \textbf{96.9} & \textbf{9.8} & \textbf{66.3} & 74.7 \\
\bottomrule
\end{tabularx}
\end{table}

%% file: sec/tables/winnoground.tex
\begin{table}[t]
    \centering
    \small
    \caption{Results (\% accuracy) on the Winoground dataset across the text, image
and group score metrics. \methodname is the only CIT model to outperform random chance on the group score metric.}
    \label{tab:winnoground}
    \begin{tabularx}{\linewidth}{Xccc}
    \toprule
    \textbf{Model} & \textbf{Text} & \textbf{Image} & \textbf{Group} \\
    \midrule
    MTurk Human & 89.50 & 88.50 & 85.50 \\
    Random Chance & 25.00 & 25.00 & 16.67 \\
    \midrule
    VinVL & 37.75 & 17.75 & 14.50 \\
    CLIP (ViT-B/32) & 30.75 & 10.50 & 8.00 \\
    SigLIP (ViT-so/14, 384) & 36.50 & 15.75 & 12.25 \\
    SigLIP 2 (ViT-so/14) & 38.25 & 19.00 & 16.00 \\
    SigLIP 2 (ViT-g/14) & 38.75 & 17.25 & 14.00 \\
    \midrule
    \methodname (ViT-B/14) & 37.50 & 16.25 & 11.25 \\
    \methodname (ViT-So/14, 384) & 42.25 & \textbf{20.50} & 17.75  \\
    \methodname (ViT-G/16, 384) & \textbf{42.50} & 20.00 & \textbf{18.50}  \\
    \bottomrule
    \end{tabularx}
\end{table}

%% file: sec/tables/blink.tex
\begin{table*}[t]
    \centering
    \scriptsize
    \caption{Results (\% accuracy) on the BLINK benchmark. TULIP demonstrates strong results across all categories, particularly excelling in vision-driven tasks, outperforming GPT-4o in some cases.}
    \label{tab:blink}
    \begin{tabularx}{\textwidth}{lCCCCCCCCCCCCCCC}
        \toprule
        \textbf{Model} & \textbf{Overall} & \textbf{Sim.} & \textbf{Count} & \textbf{Depth} & \textbf{Jigsaw} & \textbf{Art} & \textbf{Fun.-Corr.} & \textbf{Sem.-Corr.} & \textbf{Spatial} & \textbf{Local.} & \textbf{Vis.-Corr.} & \textbf{Multi-view} & \textbf{Reflect.} & \textbf{Forensic} & \textbf{IQ} \\
        \midrule
        Human & 95.67 & 96.70 & 93.75 & 99.19 & 99.00 & 95.30 & 80.77 & 96.07 & 98.25 & 98.00 & 99.42 & 92.48 & 95.14 & 100.00 & 80.00 \\
        Random Choice & 38.09 & 50 & 25 & 50 & 50 & 50 & 25 & 25 & 50 & 50 & 25 & 50 & 33.33 & 25 & 25 \\
        \midrule
        GPT-4o & 60.04 & 72.59 & 49.17 & 74.19 & 55.33 & 82.91 & 40.77 & 53.96 & 69.23 & 59.84 & 75.00 & 59.40 & 37.31 & 79.55 & 31.33 \\
        GPT-4 Turbo & 54.61 & 80.74 & 57.50 & 66.13 & 69.33 & 79.49 & 24.62 & 30.94 & 69.23 & 52.46 & 52.33 & 52.63 & 32.84 & 63.64 & 32.67 \\
        GPT-4V  & 51.14 & 78.52 & 60.83 & 59.68 & 70.00 & 79.49 & 26.15 & 28.78 & 72.73 & 54.92 & 33.72 & 55.64 & 38.81 & 34.09 & 22.67 \\
        \midrule
        LLaVA 1.6 34B & 46.80&48.89&66.67&67.74&54.67&43.59&20.77&23.74&74.83&59.02&30.81&	\textbf{62.41}&31.34&44.70&26.00 \\
        QwenVL-Max & 40.28&51.11&\textbf{56.67}&58.06&4.67&38.46&\textbf{28.46}&23.02&69.93&48.36&31.40&	51.88&\textbf{36.57}&43.94&21.33 \\ 
        Llama-3.2-11B \\
        $\quad+$ SigLIP (So/14) & 48.70 & 65.29 & 55.04 & 63.56 & 53.97 & 66.09 & 25.16 & 24.93 & 74.56 & 57.64 & 47.90 & 40.14 & 34.78 & 46.29 & 26.03\\
        $\quad+$ DINOv2 (L/16) & 49.51 & 67.13 & 53.49 & 64.08 & 56.26 & 67.88 & 23.12 & 27.59 & 75.01 & 58.21 & 46.23 & 44.66 & 33.01 & 48.56 & 28.08 \\
        \textbf{$\quad+$ \methodname (So/14)} & \textbf{50.83} & \textbf{68.29} & 55.34 & \textbf{64.29} & \textbf{57.26} & \textbf{68.39} & 25.61 & \textbf{29.61} & \textbf{76.23} & \textbf{60.01} & \textbf{48.97} & 44.96 & 35.21 & \textbf{49.07} & \textbf{28.38} \\
        \bottomrule
    \end{tabularx}
    
\end{table*}

%% file: sec/tables/mmvp.tex
\begin{table}[t]
\centering
\small
\caption{Llama-3.2 11B finetuned with several vision models on the MMVP and LLaVA benchmarks. While the LLaVA bench performance is limited by the LLM/training architecture, the MMVP benchmark shows reliance on visual representation quality.}
\label{tab:vllm}
\begin{tabularx}{\linewidth}{Xcc}
\toprule
\textbf{Model} & \textbf{MMVP} &  \textbf{LLaVA} \\
\midrule 
DINOv2 (ViT-L/16) & 16.2 & 68.5 \\
OpenAI CLIP (ViT-B/16)      & 4.5 & 80.1 \\
\midrule
SigLIP (Vit-So/14)     & 5.9 & 81.1  \\
\quad+I/I \& T/T Constrastive Learning   & 17.4 (+11.5) & 82.3 \\
\quad\quad   + Reconstruction & 18.2 (+1.2) & 82.1 \\
\quad\quad\quad   + \genaugname  (\textbf{\methodname}) &  20.3 (+2.1) & 81.9  \\
\midrule
SigLIP (Vit-B/14)     & 5.2 & 80.1  \\
\quad+I/I \& T/T Constrastive Learning      & 14.4 (+9.2) & 81.3 \\
\quad\quad   + Reconstruction & 15.8 (+1.4) & 80.8 \\
\quad\quad\quad   + \genaugname (\textbf{\methodname}) &  17.1 (+1.4) & 81.7  \\
\bottomrule
\end{tabularx}
\end{table}

%% file: sec/5_conclusion.tex
\section{Conclusion}

This work introduces \methodname, a family of multimodal self-supervised image-text contrastive foundation models that leverage learning fine-grained visual features while maintaining global semantic alignment. By unifying image-image contrastive learning with multimodal generative data augmentation, \methodname achieves SOTA performance across a range of benchmarks at scales up to 1B parameters. \methodname only represents the beginning for multi-view and generative-view models. As multimodal systems continue to advance, future work can explore broader modality integration and more efficient scaling techniques to push the boundaries of vision-language understanding.

\section*{Acknowledgements}
Authors, as part of their affiliation with UC Berkeley, were supported in part by the National Science Foundation, US Department of Defense, and/or the Berkeley Artificial Intelligence Research (BAIR) industrial alliance program.

%% file: sec/6_appendix.tex
\section*{Appendix}

The appendix consists of the following further discussion:

\begin{itemize}
    \item \autoref{app:release} discusses the model release.
    \item \autoref{app:training_data} discusses the datasets that we use for pre-training both \methodname and \genaugname.
    \item \autoref{app:datasets} discusses the datasets that we use to evalaute \methodname.
    \item \autoref{app:gecov} discusses the implementation details for the generative data augmentation portion of our approach.
    \item \autoref{app:model_config} discusses some model detail configurations.
    \item \autoref{app:vis} provides some visualizations of the self-attention weights of the \methodname model. 
\end{itemize}

\section{Code Release}
\label{app:release}

For more information on the code, and for all models, see \url{https://tulip-berkeley.github.io}.

\section{Training Data}
\label{app:training_data}

We pre-train all models on the DataComp-1B dataset \cite{gadre2023datacomp}. DataComp-1B is a large-scale dataset comprising approximately 1.4 billion image-text pairs, curated from the CommonPool collection of 12.8 billion samples. We also train with captions from Recap-DataComp-1B \cite{li2024if}, a large-scale dataset where approximately 1.3 billion images from DataComp-1B have been re-captioned using LLaMA-3-powered LLaVA-1.5. The goal of this recaptioning process is to enhance the textual descriptions associated with web-crawled image-text pairs, addressing issues like misalignment, brevity, and lack of descriptive detail in original captions. The new dataset has longer and more diverse textual annotations, increasing from an average 10.22 words per caption to 49.43 words, capturing richer contextual details.\\

\noindent\genaugname is fine-tuned on standard augmentations, as well as the following data:

\paragraph{WebVid-10M:} WebVid-10M \cite{Bain21} is a large-scale video-text dataset designed to support video-language model training and text-to-video retrieval tasks. The dataset is automatically collected from the web using a pipeline similar to Conceptual Captions \cite{sharma2018conceptual}, ensuring diverse and naturally occurring video-caption pairs. A key feature of WebVid-10M is that it focuses on real-world, diverse, and multimodal video content, making it a more challenging and representative dataset compared to traditional manually annotated datasets. The dataset spans a wide range of video types, including people performing actions, nature scenes, travel vlogs, and instructional content. Unlike other large-scale video datasets such as HowTo100M, which rely on automated speech recognition (ASR) transcriptions (often introducing noise and weak supervision), WebVid-10M provides directly associated textual descriptions, resulting in higher-quality supervision for training vision-language models.

\paragraph{MVImgNet:} MVImgNet \cite{yu2023mvimgnet} is a large-scale dataset of multi-view images, designed as a bridge between 2D and 3D vision by capturing real-world objects from multiple viewpoints. The dataset consists of 6.5 million frames extracted from 219,188 videos, covering 238 object classes with extensive annotations including object masks, camera parameters, and point clouds. Unlike single-image datasets like ImageNet, MVImgNet is built from videos, capturing objects from different angles, which naturally introduces 3D-aware visual signals.

\section{Evaluation Datasets}
\label{app:datasets}

\paragraph{ImageNet-1K:} The ImageNet-1K dataset \cite{deng2009imagenet} is a large-scale benchmark dataset widely used for training and evaluating deep learning models in computer vision. It consists of approximately 1.28 million training images, 50,000 validation images, and 100,000 test images, categorized into 1,000 distinct object classes. These classes span a diverse range of objects, including animals, vehicles, tools, and everyday items, making it a comprehensive dataset for image classification tasks. ImageNet-V2 \cite{recht2019imagenet} is a re-evaluated version of the original ImageNet dataset, designed to assess the generalization ability of models trained on ImageNet-1K. It consists of 10,000 images curated using the same class distribution and data collection process as the original validation set but sourced independently to reduce potential dataset biases. ImageNet-ReaL \cite{beyer2020we} is a re-annotated version of the ImageNet validation set, created to provide more accurate and comprehensive labels. Unlike the original ImageNet-1K validation set, where each image is assigned a single ground truth label, ImageNet-ReaL introduces multi-label annotations, acknowledging that many images contain multiple valid object categories. 

\paragraph{ObjectNet:} ObjectNet \cite{barbu2019objectnet} is a real-world test dataset designed to evaluate the robustness and generalization of image classification models beyond standard benchmarks like ImageNet-1K. It consists of 50,000 images featuring objects from 313 categories, many of which overlap with ImageNet classes. Unlike ImageNet, ObjectNet introduces systematic variations in object orientation, background, and viewpoint, making it significantly more challenging for models.

\paragraph{iNaturalist-2018:} iNaturalist-2018 \cite{van2018inaturalist} is a large-scale image classification dataset focused on fine-grained species recognition, designed to challenge models with real-world biodiversity data. It contains 437,513 training images and 24,426 validation images across 8,142 species, spanning diverse categories such as plants, insects, birds, mammals, and fungi. Unlike datasets like ImageNet, iNaturalist-2018 exhibits long-tailed class distributions, meaning some species have thousands of images while others have only a few, mimicking real-world imbalances in biodiversity data.

\paragraph{CIFAR-100:} CIFAR-100 \cite{krizhevsky2009learning} is a small-scale image classification dataset designed for evaluating machine learning models, particularly in the context of deep learning. It consists of 60,000 color images of size 32×32 pixels, with 50,000 training images and 10,000 test images. The dataset contains 100 classes, each with 600 images, and these classes are further grouped into 20 superclasses (e.g., aquatic mammals, vehicles, flowers).

\paragraph{RxRx1:} RxRx1 \cite{sypetkowski2023rxrx1} is a biological image dataset designed for evaluating domain generalization in deep learning models, specifically in the context of cellular microscopy images. It consists of 125,510 images of human cells treated with various chemical perturbations, captured using high-throughput fluorescence microscopy. A key challenge in RxRx1 is that images come from multiple experimental batches across four cell types, introducing batch effects—systematic variations that can hinder model generalization. 

\paragraph{fMoW:} fMoW (Functional Map of the World) \cite{christie2018functional} is a large-scale remote sensing dataset designed to evaluate model performance on satellite image classification and change detection tasks. It contains over 1 million images from diverse geographic locations, covering 62 categories of functional land use and infrastructure, such as airports, military facilities, bridges, and solar farms. The dataset includes images captured under varied lighting conditions, seasonal changes, and resolutions, making it a challenging benchmark for real-world geospatial analysis. 

\paragraph{Infographic:} InfographicVQA \cite{mathew2022infographicvqa} is a dataset designed for Visual Question Answering (VQA) on infographics, which are complex document images combining text, graphics, and data visualizations. The dataset consists of 5,485 images and 30,035 questions, with annotations requiring reasoning over various elements such as tables, figures, maps, and textual content. Unlike traditional VQA datasets, InfographicVQA places emphasis on elementary reasoning skills, including counting, sorting, and basic arithmetic operations.

\paragraph{Winnoground:} Winoground \cite{thrush2022winoground} is a dataset introduced to evaluate the ability of vision-and-language models to perform visio-linguistic compositional reasoning. Each of the 400 examples in the dataset consists of two images and two captions, where both captions contain the same set of words arranged differently, leading to distinct meanings. The task requires models to correctly match each image with its corresponding caption, testing their understanding of how word order affects meaning in a visual context.

\begin{figure}
    \centering
    \includegraphics[width=\linewidth]{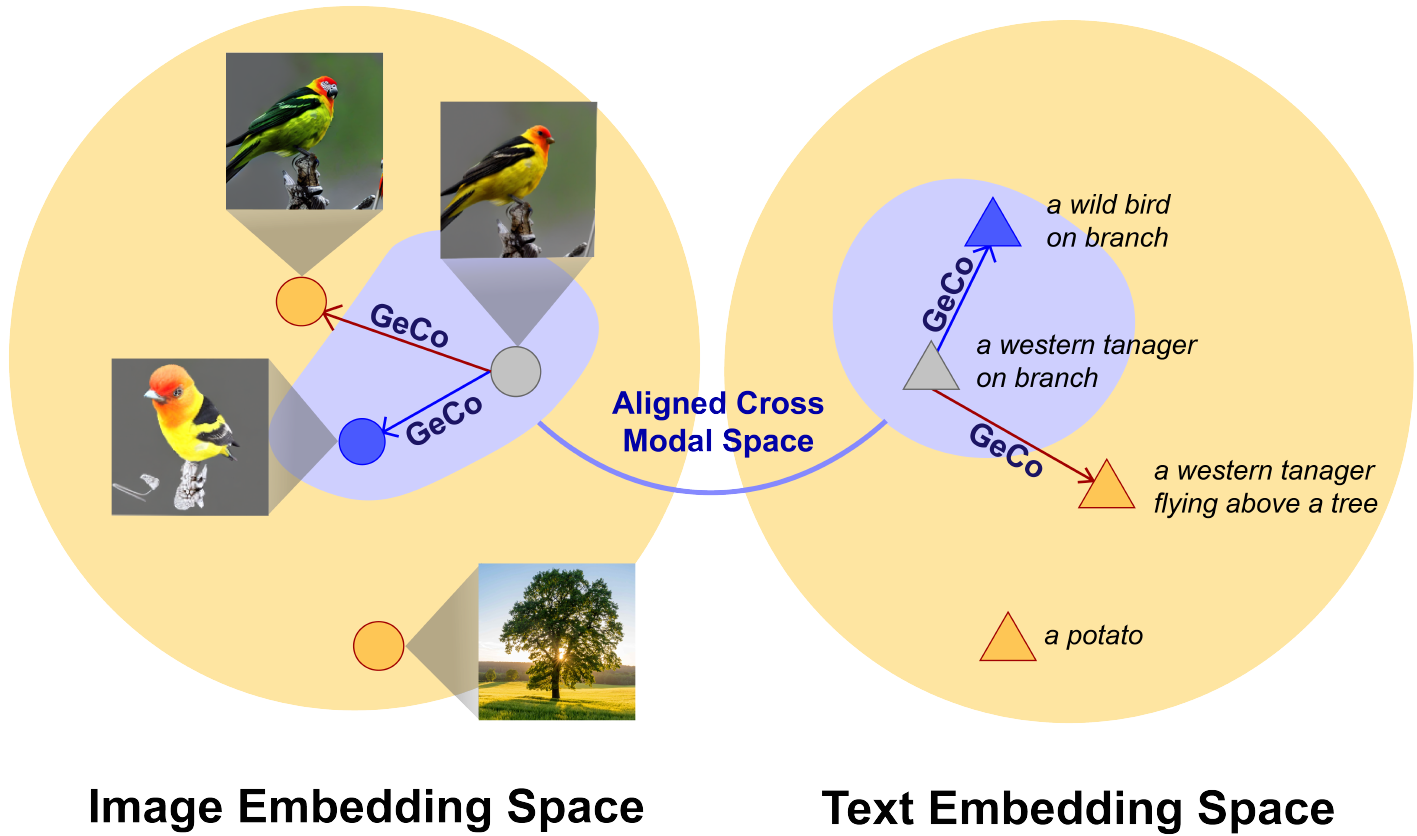}
    \vspace{3pt}
    \caption{(Top) \genaugname generates positive (in blue region) and hard negative augmentations (in yellow region) of both images and text. Hard negative is closer to the `positive region' while randomly sampled images or text are further.}
    \label{fig:gecov_example_2}
\end{figure}

\section{Data Augmentation}
\label{app:gecov}

As discussed in \autoref{ssec:generative_transformations}, we generate both positive view and negative views for contrastive learning. We show some example in \autoref{fig:gecov_example}. To generate positive view of the image, we input positive embedding $E_p$. and a high image classifier free guidance (cfg) scale 5. To generate negative view of the image, we input negative embedding $E_n$ to the model with a lower image cfg scale 3. To generate paraphrases for the image augmentation model, we use the prompt in \autoref{fig:gecov_prompt}, which can generate a positive and a negative example for an input caption. \autoref{fig:gecov_example_2} gives additional insight into our data augmentation method.

\begin{figure*} 
\begin{minipage}{\linewidth}
\begin{tcolorbox}[colback=gray!10, colframe=black]
Given an input caption describing an image, generate two variants:\\

Positive Example: A paraphrased version that preserves the exact meaning using synonyms, grammatical reordering, or structural changes (e.g., active/passive voice).\\

Negative Example: A minimal, plausible alteration that subtly contradicts the original meaning. Prioritize compositional changes (e.g., swapped roles, spatial relations, object attributes, or verb actions) while keeping lexical overlap high. The negative should be visually distinct but textually similar to trick models.\\

Guidelines:
Positive Paraphrase:

Use synonyms (``cube" → ``square"), reorder clauses (``X beside Y" → ``Y next to X"), or adjust syntax (``holding a leash" → ``gripping a dog’s lead").

Ensure no key details (objects, relationships, attributes) are altered.

Hard Negative:

Swap Roles/Relations: Invert subject-object relationships (``a man riding a horse" → ``a horse beside a man").

Modify Prepositions/Spatial Logic: Change directional/positional cues (``left of" → ``under").

Alter Attributes: Adjust colors, sizes, or quantities (``three red apples" → ``two green apples").

Reorder Phrases with Identical Words: Use the same words in a different order to invert meaning (``plants surrounding a lightbulb" → ``a lightbulb surrounding some plants").\\

Example:
Input:
``A chef in a white hat is slicing vegetables on a stainless steel counter while a cat watches from the windowsill."\\

Positive:
``A cook wearing a white cap chops veggies on a shiny metal countertop as a feline observes from the window ledge."
(Synonym substitution + rephrasing)\\

Negative:
``A cat in a white hat is slicing vegetables on a stainless steel counter while a chef watches from the windowsill."
(Role swap: ``chef" $\xrightarrowdbl{}$ ``cat" + retained details create a contradictory but plausible scene.)

\end{tcolorbox}
\end{minipage}
\caption{The \genaugname prompt.}
\label{fig:gecov_prompt}
\end{figure*}

\section{Model Configurations}
\label{app:model_config}

\autoref{tab:model_config} provides an overview of our model configurations, detailing key parameters such as image size, sequence length, hidden size, number of layers, and text context length. We follow SigLIP 2 to use So400M language encoder for ViT-G/16.

\begin{table*}\centering
    \renewcommand{\arraystretch}{1.22}
    \begin{tabularx}{0.7\textwidth}{lXXXX}
        \toprule
        Hyperparameter & ViT-G/16 & ViT-SO400M & ViT-H-14 & ViT-B-16 \\
        \midrule
        Embed Dim & 1536 & 1152 & 1152 & 768 \\
        Init Logit Bias & -10 & -10 & -10 & -10 \\
        Image Size & 384 & 384 & 224 & 224 \\
        Patch Size & 16 & 14 & 14 & 16 \\
        Layers (Vision) & 43 & 27 & 32 & 12 \\
        Width (Vision) & 1536768 & 1152768 & 1280 & 768 \\
        Head Width (Vision) & 64 & 64 & 80 & 64 \\
        MLP Ratio & 3.7362 & 3.7362 & 3.7362 & 4.0 \\
        Pooling & map & map & tok & map \\
        Projection & none & none & linear & none \\
        Context Length & 70 & 70 & 70 & 70 \\
        Vocab Size & 109871 & 109871 & 109871 & 109871 \\
        Tokenizer & tulip-tokenizer & tulip-tokenizer & tulip-tokenizer & tulip-tokenizer \\
        Width (Text) & 1152 & 1152 & 1024 & 768 \\
        Heads & 16 & 16 & 16 & 12 \\
        Layers (Text) & 27 & 27 & 24 & 12 \\
        No Causal Mask & True & True & True & True \\
        Projection Bias & True & True & True & True \\
        Pool Type & last & last & last & last \\
        Norm Eps & $10^{-6}$ & $10^{-6}$ & $10^{-6}$ & $10^{-6}$ \\
        Activation Approx. & tanh & tanh & tanh & - \\
        Attentional Pool & False & False & False & False \\
        Attn Pooler Queries & 256 & 256 & 256 & 256 \\
        Attn Pooler Heads & 8 & 8 & 8 & 8 \\
        Pos Embed Type & learnable & learnable & learnable & learnable \\
        Final LN After Pool & False & False & False & False \\
        Output Tokens & False & False & False & False \\
        Timm Pool & map & map & avg & map \\
        Timm Proj & none & none & linear & none \\
        Timm Proj Bias & False & False & False & False \\
        Timm Drop & 0.0 & 0.0 & 0.0 & 0.0 \\
        Timm Drop Path & None & None & None & None \\
        \bottomrule
    \end{tabularx}
    \caption{Comparison of Vision Transformer (ViT) Model Hyperparameters for different \methodname variants.}
    \label{tab:model_config}
\end{table*}

\section{Attention Visualization}
\label{app:vis}

\begin{figure*}
  \centering
  \includegraphics[width=0.7\linewidth]{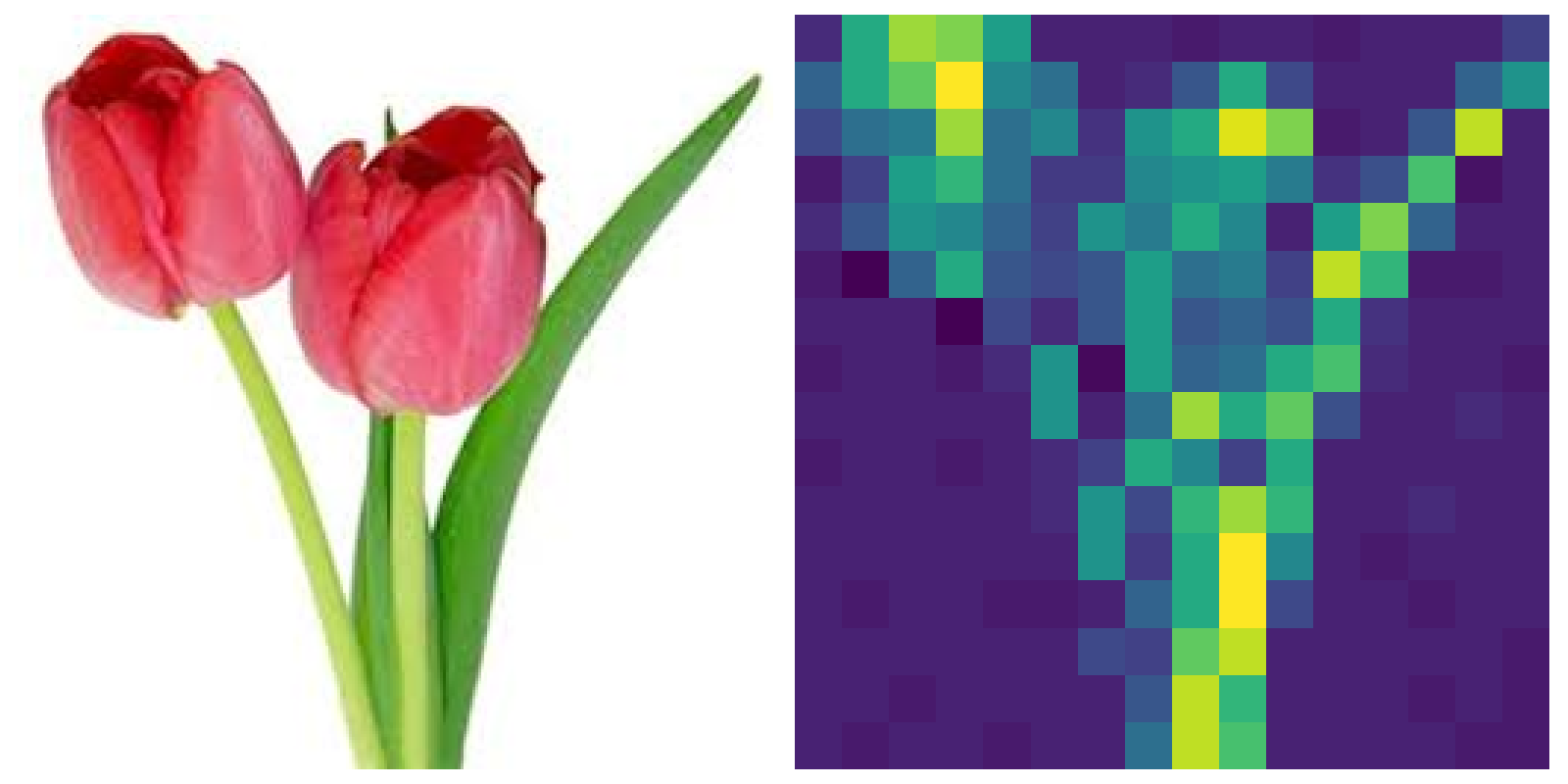}
  \vspace{0.5cm} %
  \includegraphics[width=0.7\linewidth]{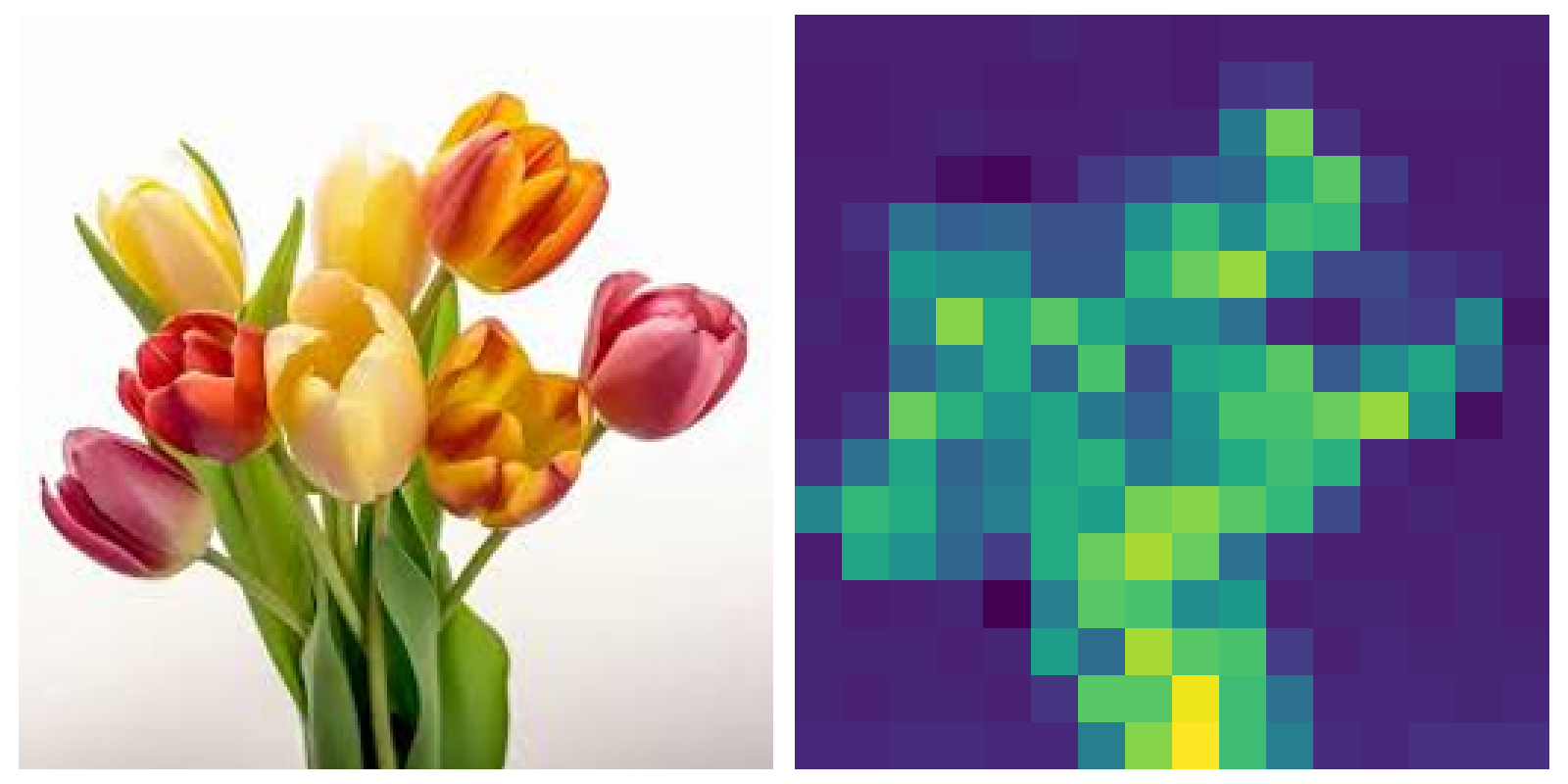}
  \vspace{0.5cm} 
  \includegraphics[width=0.7\linewidth]{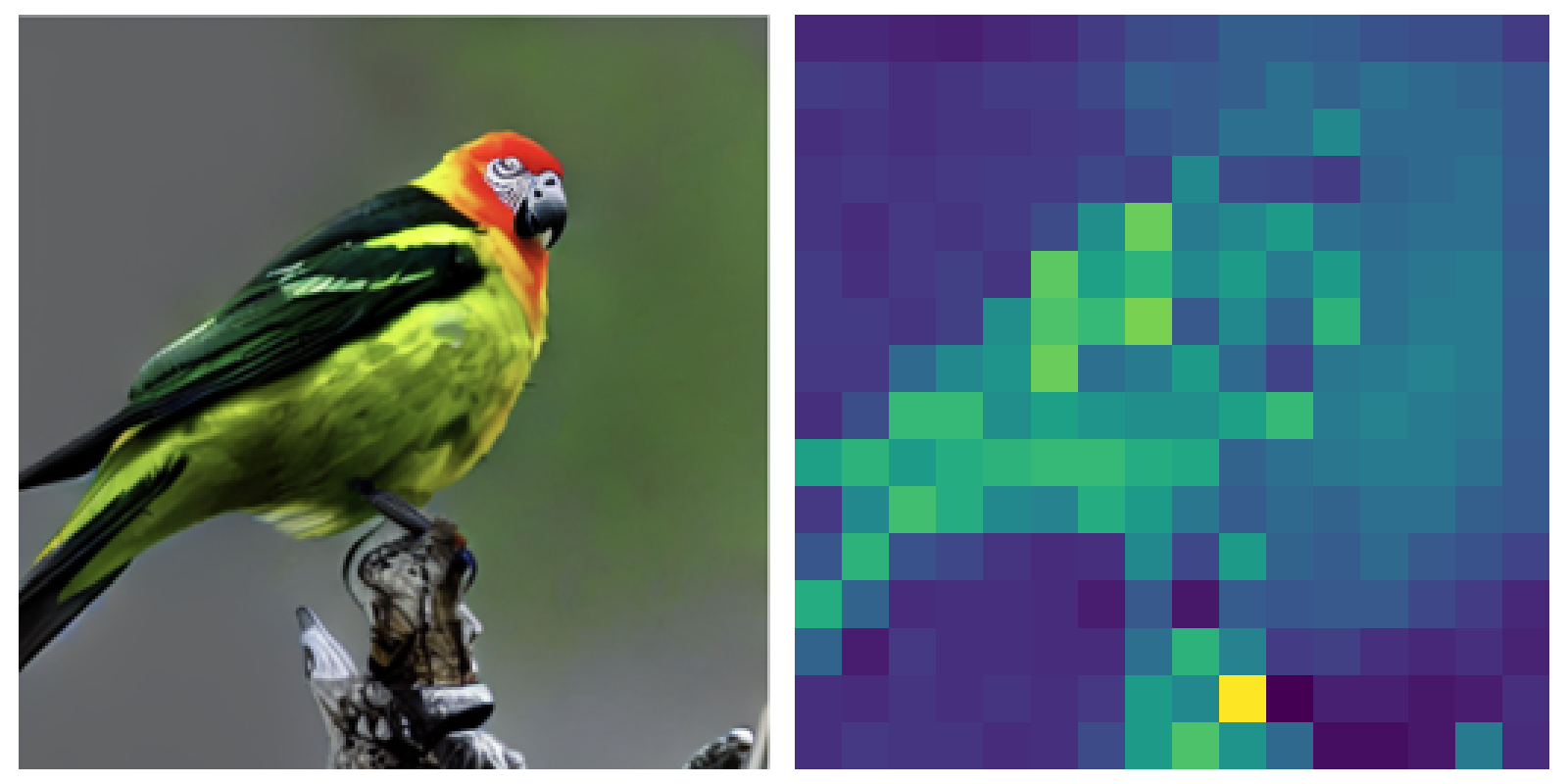}

  \caption{Visualization of the attention heads. Attention maps are averaged across transformer blocks, then up-sampled to the resolution of the original image.}
  \label{fig:attention heatmaps}
\end{figure*}

\autoref{fig:attention heatmaps} shows a visualization of the attention heads of the So/14 model. We can see that similar to DINOv2, the model performs local semantic segmentation as an emergent behavior the in the attention weights.

